\documentclass[lettersize,journal]{IEEEtran}
\usepackage{amsmath,amssymb,amsfonts,amsthm}
\usepackage{algorithm,algpseudocode}
\usepackage{booktabs,multirow,array}
\usepackage{graphicx}
\graphicspath{{figs/}}
\usepackage{textcomp,url}
\usepackage{cite}

\newtheorem{definition}{Definition}

\newtheorem{proposition}{Proposition}
\newtheorem{assumption}{Assumption}
\usepackage{pdfpages}
\usepackage[hidelinks]{hyperref}
\begin{document}
\title{Learned Network Dynamics for Model Predictive \\ Traffic Signal Control under Incidents}
\author{Dang~Viet~Anh~Nguyen, Carlos~Lima~Azevedo, Tomer~Toledo, and Filipe~Rodrigues%
\thanks{This work has been submitted to the IEEE for possible publication. Copyright may be transferred without notice, after which this version may no longer be accessible.}%
\thanks{D. V. A. Nguyen, C. Lima Azevedo and F. Rodrigues are with the Department of Technology, Management and Economics, Technical University of Denmark, 2800 Kongens Lyngby, Denmark.}%
\thanks{T. Toledo is with the Faculty of Civil and Environmental Engineering, Technion -- Israel Institute of Technology, Haifa 32000, Israel.}}
\markboth{Submitted to IEEE Transactions on Control Systems Technology}{Nguyen \MakeLowercase{\textit{et al.}}: Learned Network Dynamics for Model Predictive Traffic Signal Control}
\maketitle
\begin{abstract}
Model predictive control (MPC) of urban traffic signals relies on a model of how queues respond to signal decisions, which is hard to specify and calibrate in heterogeneous networks, particularly when incidents change capacities and turning shares. Measurements add a second limitation: the controlled lanes at the stop line are often short, so queues extending upstream of them fall outside lane-level measurements. We propose GNN-MPC, a receding-horizon controller that plans with action-conditioned network dynamics learned by graph neural networks. Phase sequences are coordinated across intersections by approximate best responses, with an acceptance test that keeps the predicted cost no higher than that of the initial plan. In microscopic simulation under non-recurrent incidents, GNN-MPC reduces travel time and increases completed trips relative to all tested controllers, on a homogeneous grid and a heterogeneous 21-intersection network. Controlled comparisons on the heterogeneous network distinguish the roles of observation and prediction. With nearly as many vehicles entering the network, GNN-MPC reduces travel time by 13\% relative to MaxPressure with the same upstream measurement, and GNN-MPC without the upstream representation does not reproduce its performance. Under the same traffic measurements, planner and objective, replacing the learned dynamics with calibrated store-and-forward predictors increases travel time by 31--34\%. Observing hidden demand and predicting its response to control are thus complementary requirements of network-level predictive control.
\end{abstract}
\section{Introduction}

\IEEEPARstart{N}{on-recurrent} incidents, such as crashes and stalled vehicles, reduce the capacity of road links without warning and are a major source of traffic delay~\cite{knoop2008capacity}. In urban networks, an incident changes how queues form, discharge and propagate, so that signal settings suited to earlier conditions may become less effective; this motivates repeated, state-dependent optimization by model predictive control (MPC). The effectiveness of MPC depends on the information available about the system state and on the fidelity with which its model predicts the consequences of candidate actions~\cite{mayne2000constrained,hewing2020learning}. In networked systems such as urban signal control~\cite{de1998optimal,ye2019survey}, measurements often cover only part of the state, and incidents place both requirements under additional strain.

\begin{figure*}[t]
    \centering
    \includegraphics[width=0.78\textwidth]{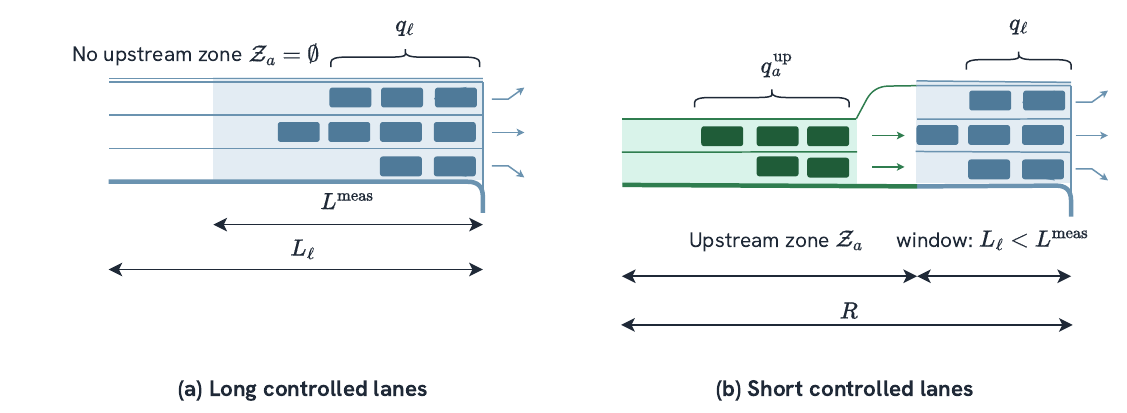}
    \caption{Observation of an approach. Each lane window measures the queue $q_\ell$ on a controlled lane, from the stop line over at most the detection range $L^{\mathrm{meas}}$. (a)~Long controlled lanes ($L_\ell \geq L^{\mathrm{meas}}$): the windows cover the downstream part of the approach. (b)~Short controlled lanes ($L_\ell < L^{\mathrm{meas}}$): each window ends where its lane begins, and vehicles queued upstream of the lanes are not measured; an upstream zone $\mathcal{Z}_a$ can count these stopped vehicles, $q_a^{\mathrm{up}}$, but not the movements they will take or their future arrivals.}
    \label{fig:coverage}
\end{figure*}

Signal controllers, from MaxPressure~\cite{varaiya2013max} to learning-based~\cite{noaeen2022reinforcement} and predictive controllers~\cite{ye2019survey}, commonly represent the traffic state by the queues and vehicle counts on the lanes at the stop line, measured by detectors or video detection zones over a limited range. Queues can extend beyond this range, and on urban approaches this is frequent for a geometric reason: approaches often widen into turning bays near the intersection, so that the \emph{controlled lanes}, i.e., the lanes at the stop line, each assigned to specific movements, can be much shorter than the detection range. A lane-level measurement then covers the lane fully but ends where the lane begins, so that a queue extending upstream of it is not measured: with long controlled lanes, the measured windows cover the downstream part of the approach (Fig.~\ref{fig:coverage}a), whereas with short ones, the queue continues upstream of the windows (Fig.~\ref{fig:coverage}b). A controller that minimizes the measured queues may then leave demand waiting outside the observed region. This limitation is not specific to disruptions, but incidents can make it more consequential: when a blockage reduces the discharge of an approach or of the link it feeds, queues may accumulate farther upstream, beyond the controlled lanes.

For predictive control, this setting presents two coupled challenges. The first is observation coverage: the measurements may miss part of a queue, and an additional approach-level measurement, from upstream detectors or video-based queue detection, can count the vehicles queued upstream of the controlled lanes but cannot assign them to movements, since the lanes they will enter have not yet opened where they wait. The second is prediction-model fidelity: the controller must predict how alternative signal actions affect the inflow into the approaches and their controlled lanes, the discharge and the propagation of queues across the network, and how disruptions affecting discharge, routing and spillback alter this evolution. Traffic flow models, such as store-and-forward, cell-transmission and point-queue formulations, represent these processes through specified and calibrated flow, capacity and routing relationships~\cite{aboudolas2009store,daganzo1994cell,lin2023network}. They can include incident-dependent capacities and spillback, but are challenging to specify and calibrate in heterogeneous networks, and each incident must be translated into model parameters, such as reduced capacities and altered turning shares. Learned transition models instead estimate these relationships from recorded traffic states, signal actions and observed incident descriptors, so that the effect of an incident on the dynamics is learned from data rather than specified, provided that their predictions are conditioned on candidate control decisions.

Learning-based MPC provides a framework for incorporating such models into predictive control~\cite{hewing2020learning}, and several signal controllers plan with learned traffic models (Section~\ref{sec2:related}). Model-free reinforcement-learning methods instead learn a policy, a direct mapping from observed states to signal actions, by trial and error in simulation, without predicting how actions affect traffic~\cite{noaeen2022reinforcement}; their performance under incidents varies across architectures and incident conditions~\cite{nguyen2025robustness}. However, existing evaluations provide limited evidence on the relative roles of observation coverage and action-conditioned prediction under disruptions. When controllers differ in both their measurements and predictive models, performance improvements cannot be attributed to either without appropriately controlled comparisons.

We address this problem with GNN-MPC, a receding-horizon signal controller that plans with action-conditioned network dynamics learned by graph neural networks (GNNs) at two levels: a network-level model captures interactions between intersections, and a lane-level model predicts the quantities in the control objective, including the demand upstream of the controlled lanes. The phase sequence of each intersection is constructed by probabilistic beam search, and the sequences are coordinated across intersections through sequential best responses. The model is trained offline, while the controller replans at each control step using updated traffic measurements and, when available, detected incident information. We evaluate GNN-MPC under non-recurrent incidents and use controlled observation and predictor comparisons to investigate the sources of its performance differences.

The main contributions are:
\begin{itemize}
    \item We develop an action-conditioned, two-level learned dynamics model that predicts network- and lane-level traffic under candidate signal decisions, incorporating upstream queue measurements and detected incident conditions.

    \item We integrate the learned model into coordinated receding-horizon control using probabilistic beam search and sequential best responses, with an acceptance rule that guarantees a non-increasing predicted cost relative to the initial plan.

    \item On a homogeneous grid and a heterogeneous 21-intersection network under non-recurrent incidents, GNN-MPC achieves the lowest average travel time in every incident scenario and the highest mean throughput against Fixed-Time, MaxPressure and five reinforcement-learning controllers.

    \item Through controlled comparisons with upstream-aware controller variants and analytical predictor substitutions under matched measurements and planning, we investigate the complementary roles of observation coverage and predictive-model choice. They indicate that representing upstream demand changes how much of the scheduled demand enters the network, whereas the choice of predictive model also affects how much of it completes its trips.
\end{itemize}

Section~\ref{sec2:related} reviews related work, Sections~\ref{sec3:gnn-mpc}--\ref{sec:planning} present the problem formulation and the method, and Sections~\ref{sec4:experiment}--\ref{sec:conclusion} the experimental design, results and conclusions.

\section{Related Work}
\label{sec2:related}

Predictive signal control with analytical traffic flow models has been developed extensively since the early work of De Schutter and De Moor~\cite{de1998optimal,ye2019survey}. Its traffic models range from store-and-forward formulations underlying network-wide strategies such as TUC~\cite{diakaki2002multivariable,aboudolas2009store}, including extensions with link travel times~\cite{lin2011fast}, to cell-transmission~\cite{daganzo1994cell} and lane-level point-queue models~\cite{lin2023network}. Distributed formulations decompose the network-wide problem across intersections~\cite{ye2015distributed,mohebifard2018distributed}, and SelectLight combines point-queue prediction with a learned policy that selects among locally generated signal plans~\cite{luo2026selectlightlearningselectsignal}. These approaches rely on explicitly specified flow, capacity and routing relationships. On the observation side, queues beyond stop-line detection can be estimated from upstream measurements and shockwave models~\cite{liu2009queue}, lane-level estimation must infer how arrivals are distributed across lanes~\cite{lee2015lane}, the state representation and sensing range affect learned controllers~\cite{zhang2022expression}, and capacity-aware MaxPressure variants account for limited storage and spillback~\cite{varaiya2013max,gregoire2015capacity}. These studies address incomplete observation and storage constraints, but not how the benefit of upstream observation depends on the model used to predict the consequences of candidate signal actions.

Learning-based MPC incorporates data-driven models into receding-horizon optimization~\cite{hewing2020learning}. In signal control, learned models have been used for MPC at an isolated intersection~\cite{tang2024neuralmpc}, for neural system identification with locally linearized optimization of green splits~\cite{yin2024dlmpc}, and to approximate a long-term throughput term of the MPC objective~\cite{zhu2026learningmpc}; other approaches plan through learned latent or discrete world models~\cite{devailly2024mujam,wang2024worldlight}, or transfer a shared dynamics model across observation configurations~\cite{huang2026amm}. Model-free reinforcement learning instead learns the control policy itself, with various mechanisms for coordinating intersections~\cite{noaeen2022reinforcement,ruan2024coslight}, and graph-based traffic forecasters predict future states from historical observations without conditioning on candidate signal actions~\cite{li2018dcrnn,rahmani2023gnn}. GNN-MPC differs from these approaches in predicting, with an explicit action-conditioned graph model, the network- and lane-level quantities that enter the control objective, including the demand upstream of short controlled lanes, and in using these predictions in coordinated receding-horizon optimization over discrete phases.

\section{Problem Formulation}
\label{sec3:gnn-mpc}

We consider network-wide signal control with discrete phase decisions at intersections, based on aggregate measurements of limited spatial coverage.

\subsection{Traffic Network and Control Inputs}
\label{subsec:actions}

Consider an urban road network with a set of signalized intersections $\mathcal{V}$. For intersection $v\in\mathcal{V}$, $\mathcal{L}^v$ denotes its \emph{controlled lanes}, the lanes at the stop line, each assigned to specific movements, i.e., connections to outgoing lanes; lane $\ell$ has length $L_\ell$. Controlled lanes arriving from the same direction form an \emph{approach} $a$ with lane set $\mathcal{L}_a$, and $\mathcal{L}$ and $\mathcal{A}$ denote all controlled lanes and approaches. Each intersection has a finite set $\mathcal{U}_v$ of admissible green phases, taken from its installed signal program; a phase gives green simultaneously to a set of non-conflicting movements, so that it can serve several approaches, and an approach can be served by several phases. At every control interval $\Delta t$, the controller selects a phase $u_v(t)\in\mathcal{U}_v$ for each intersection, forming the joint input $u(t)=(u_v(t))_{v\in\mathcal{V}}$; clearance intervals are inserted when consecutive phases differ.

\subsection{Measurement Model and Observation Coverage}
\label{subsec:coverage}

The vehicle-level traffic state is not available to the controller, which observes aggregate measurements over a limited distance upstream of each stop line (Fig.~\ref{fig:coverage}). On the controlled lanes, measurements are lane-based and cover each lane upstream from the stop line up to the detection range $L^{\mathrm{meas}}$; they comprise the queue $q_\ell(t)$, i.e., the number of vehicles in the window that have stopped (speed below 0.1~m/s) since entering the window, including those moving again, the accumulated waiting time $w_\ell(t)$ of these vehicles, the incoming vehicle count and speed, and the occupancy, as obtainable from lane-specific detectors or video detection zones~\cite{klein2006detector}. When a lane is shorter than $L^{\mathrm{meas}}$, its measurement ends where the lane opens. Upstream of that point, an approach $a$ can be covered by an \emph{upstream zone} $\mathcal{Z}_a$, which extends to a distance $R$ from the stop line and is disjoint from the lane windows. Since vehicles there have not yet been assigned to their lanes, the zone is measured per approach: the \emph{upstream halt} $q_a^{\mathrm{up}}(t)$ is the number of vehicles in $\mathcal{Z}_a$ that are stopped at the decision instant. Queues beyond these regions are not measured. The construction of the zones on the real network is given in Supplementary Material, Sec.~S2.

These measurements reveal queues within the observed regions, but not the inflow into each zone, the lanes that approaching vehicles will enter, or demand beyond the coverage. A controller that minimizes only the measured lane queues may favor decisions that hold vehicles upstream of the lane windows, where they are not counted; the upstream halt makes part of this demand observable and allows it to be penalized. Demand beyond the coverage remains unobserved.

\subsection{Incident Information}
\label{subsec:incident}

\begin{assumption}[Incident information]
\label{ass:incident}
The occurrence, location, and number of blocked lanes of an incident become known to the controller when the incident starts, and its clearance becomes known when it ends. The incident duration is not known in advance.
\end{assumption}

Detection delays and localization errors are outside the scope of this study, and future incident occurrences and clearance times are not known to the controller.

\subsection{Information State and Traffic Dynamics}
\label{subsec:dynamics}
\label{sec:state_representation}
\label{sec:problem}

Since the physical traffic state is only partially observed through these measurements, the controller operates on an \emph{information state}
\begin{equation}
\label{eq:info_state}
    X(t)=\big(y(t-H+1:t),\,c(t)\big),
\end{equation}
where $y$ collects the lane and upstream measurements, $H$ is the history length, and $c(t)$ collects static network attributes, observed incident descriptors and known control information. A finite history is not assumed to be a sufficient Markov representation of the traffic system. Its evolution can be written as
\begin{equation}
\label{eq:true_dynamics}
    X(t+1)=F\big(X(t),u(t),\xi(t)\big),
\end{equation}
where $F$ denotes the unknown effective dynamics and $\xi(t)$ unobserved influences such as demand variability, driver behavior and traffic beyond the coverage. A predictive model $\hat{F}$ approximates this evolution under candidate actions,
\begin{equation}
\label{eq:surrogate}
    \hat{X}(t+1)=\hat{F}\big(X(t),u(t)\big).
\end{equation}
In multi-step prediction, the measurement history and the control information are updated with the predicted states and planned actions; incident status is held over the horizon, and elapsed-time descriptors advance deterministically.

\subsection{Network-Wide Predictive Control Problem}
\label{subsec:mpc_problem}

At each step $t$, the controller seeks a joint phase sequence $U=(u_{0|t},\ldots,u_{n_u-1|t})$ over the control horizon $n_u\leq n_y$, where $n_y$ is the prediction horizon, $u_{\tau|t}$ is the joint phase planned for step $t+\tau$ with component $u_{v,\tau|t}$ at intersection $v$, and $\hat{X}_{\tau|t}$ is the predicted information state:
\begin{align}
    \min_U\quad & J_t(U)=\sum_{\tau=1}^{n_y}\Phi\big(\hat{X}_{\tau|t}\big)
    \label{eq:mpc_obj}\\
    \text{s.t.}\quad & \hat{X}_{0|t}=X(t), \notag\\
    & \hat{X}_{\tau+1|t}=\hat{F}\big(\hat{X}_{\tau|t},u_{\tau|t}\big),\;\; \tau=0,\ldots,n_y-1,
    \label{eq:dynamics}\\
    & u_{v,\tau|t}\in\mathcal{U}_v,\;\; v\in\mathcal{V},\;\; \tau=0,\ldots,n_u-1,
    \label{eq:control_constraint}\\
    & u_{\tau|t}=u_{n_u-1|t},\;\; \tau=n_u,\ldots,n_y-1.
    \label{eq:control_hold}
\end{align}
Between the end of the control horizon and the end of the prediction horizon, the last planned input is held, so that the cost is defined over the full prediction horizon, and only the first joint phase is applied before the problem is solved again with updated observations. The stage cost is
\begin{equation}
\label{eq:stage_cost}
\Phi(\hat{X}) =
    \frac{1}{\kappa_q}
    \Bigg(
        \sum_{\ell\in\mathcal{L}}\hat{q}_\ell
        +\sum_{a\in\mathcal{A}^{\mathrm{up}}}\hat{q}_a^{\mathrm{up}}
    \Bigg)
    +\frac{\lambda}{\kappa_w}
    \sum_{\ell\in\mathcal{L}}\hat{w}_\ell,
\end{equation}
where $\hat{q}_\ell$, $\hat{q}_a^{\mathrm{up}}$ and $\hat{w}_\ell$ are the predicted lane queue, upstream halt and lane waiting time contained in $\hat{X}$, i.e., at step $t+\tau$ when $\Phi$ is evaluated at $\hat{X}_{\tau|t}$, with indices omitted for readability; $\mathcal{A}^{\mathrm{up}}\subseteq\mathcal{A}$ is the set of approaches with upstream zones, $\kappa_q,\kappa_w>0$ are reference scales estimated from training data, and $\lambda\geq0$ weights the waiting time. Each upstream queue is counted once per approach; without upstream zones and with $\lambda=0$, $\Phi$ reduces to the total predicted lane queue. The cost is thus defined on predicted measurements rather than on an estimate of the vehicle-level state, which the controller does not reconstruct. The phases of all intersections are optimized jointly against a single network-wide objective, so that each signal decision is assessed by its network-wide consequences; the approximate solution of Section~\ref{sec:planning} retains this property, since each intersection's best response minimizes the same objective. Because the phases are discrete, exhaustive optimization grows exponentially with the number of intersections and the control horizon; Sections~\ref{sec:GNNs} and~\ref{sec:planning} develop the predictive model and the coordinated procedure used to solve \eqref{eq:mpc_obj}--\eqref{eq:control_hold} approximately.

\section{Learned Network Dynamics Model}
\label{sec:GNNs}
\label{subsec:overview}

\begin{figure*}[t]
    \centering
    \includegraphics[width=0.92\textwidth]{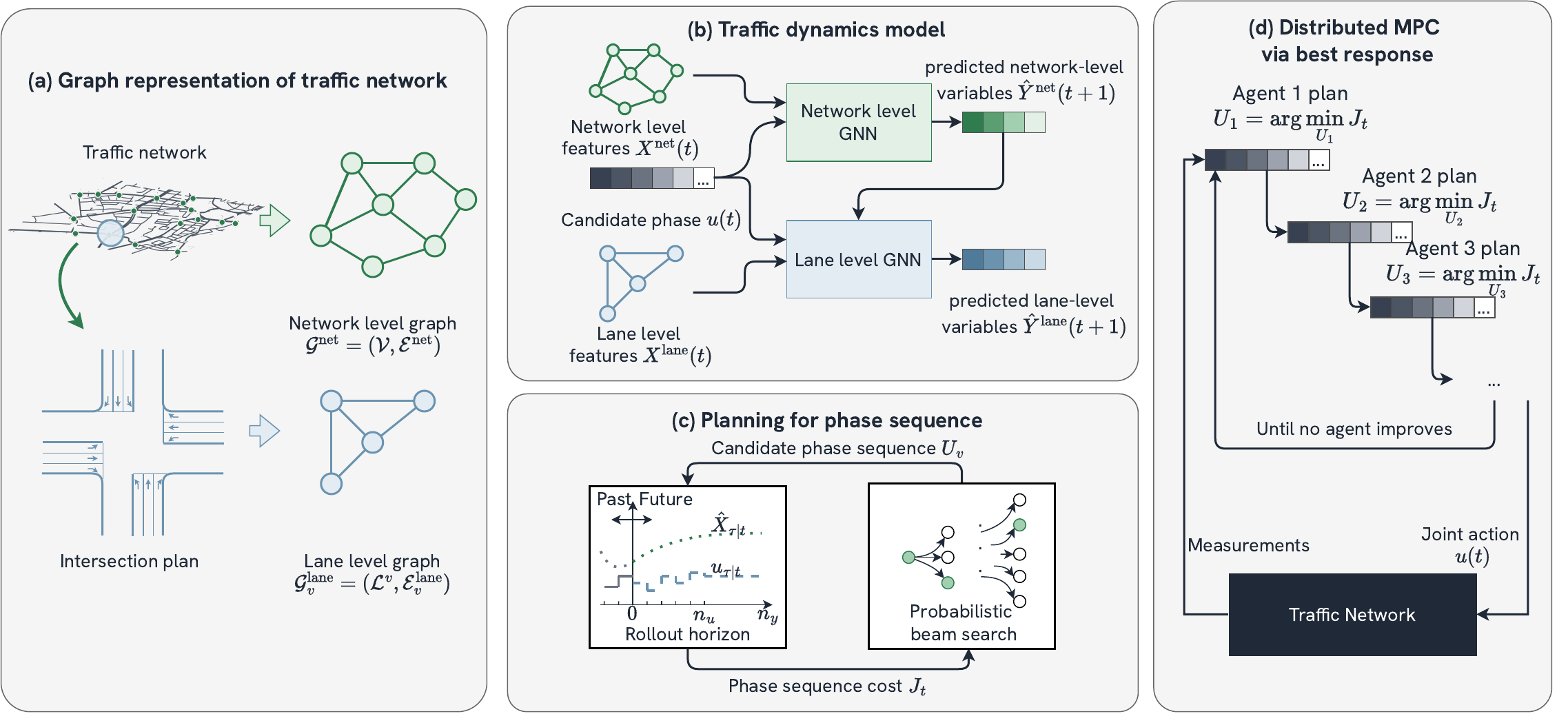}
    \caption{Overview of GNN-MPC at one control step: (a)~two-level graph representation of the network, (b)~learned network dynamics model, (c)~phase-sequence search for an individual intersection, and (d)~coordination through iterated best responses. In~(c), the predicted information state $X(\tau)$ and the planned phases $U(\tau)$ are shown over the control horizon $n_u$ and the prediction horizon $n_y$. Panels (a)--(b) provide the predictive model, while (c)--(d) constitute the coordinated planner.}
    \label{fig:framework}
\end{figure*}

GNN-MPC addresses the predictive control problem~\eqref{eq:mpc_obj}--\eqref{eq:control_hold} through two components (Fig.~\ref{fig:framework}): a learned network dynamics model $f_\theta$, which implements the predictive transition $\hat{F}$ of~\eqref{eq:surrogate} for network- and lane-level traffic quantities under candidate signal actions (Section~\ref{sec:GNNs}), and a coordinated receding-horizon planner, which evaluates alternative phase sequences by rolling out this model (Section~\ref{sec:planning}). Fig.~\ref{fig:framework} follows one control step. The network is represented by two graphs (Fig.~\ref{fig:framework}a): a network-level graph whose nodes are the signalized intersections, and, for each intersection, a lane-level graph whose nodes are its controlled lanes (Section~\ref{subsec:graph}). Given the measured features and a candidate phase, the network-level model predicts the variables of each intersection, and the lane-level model, conditioned on these predictions, predicts the lane variables, including the queues, upstream halts and waiting times of the stage cost (Fig.~\ref{fig:framework}b; Section~\ref{subsec:hier_transition}). Applied recursively, the model rolls out a candidate phase sequence over the prediction horizon, and the accumulated cost $J_t$ scores the sequence; a probabilistic beam search uses these scores to construct the phase sequence of one intersection (Fig.~\ref{fig:framework}c; Section~\ref{subsec:optimization}). The intersections then take turns computing such best responses against the current plans of the others, all minimizing the same cost $J_t$, until a full pass no longer improves the joint plan sufficiently or a pass limit is reached; the first phase of the joint plan is applied, and the procedure is repeated at the next step with new measurements (Fig.~\ref{fig:framework}d; Section~\ref{subsec:ibr}).

\subsection{Two-Level Graph Representation}
\label{subsec:graph}

The model represents traffic at two scales: the \emph{network level} captures interactions between signalized intersections, and the \emph{lane level} preserves differences among the controlled lanes of an intersection, such as their movements, discharge and exposure to incidents.

\begin{definition}[Network-level graph]
\label{def:network}
The network-level graph is a directed graph $\mathcal{G}^{\mathrm{net}}=(\mathcal{V},\mathcal{E}^{\mathrm{net}})$, where each node $v\in\mathcal{V}$ represents a signalized intersection. An edge $(v_i,v_j)\in\mathcal{E}^{\mathrm{net}}$ exists if traffic leaving $v_i$ can reach $v_j$ without passing another signalized intersection, within a path length of at most $D$.
\end{definition}

On the networks considered here, $D = 1{,}000$~m, which excludes no connection between adjacent signals on the grid (Supplementary Material, Sec.~S2).

\begin{definition}[Lane-level graph]
\label{def:intersection}
For each intersection $v\in\mathcal{V}$, the lane-level graph is $\mathcal{G}^{\mathrm{lane}}_v=(\mathcal{L}^v,\mathcal{E}^{\mathrm{lane}}_v)$, where each node represents a controlled lane and an edge between lanes $\ell_i$ and $\ell_j$ exists if they receive green simultaneously in at least one admissible phase of $\mathcal{U}_v$.
\end{definition}

The edges of $\mathcal{G}^{\mathrm{net}}$ follow the road connections through which traffic propagates between intersections, and $\mathcal{G}^{\mathrm{lane}}_v$ encodes the phase compatibility of the installed signal program, so that the lanes served by one phase form a clique. Both topologies are fixed: candidate phases condition the predictions without changing the connectivity, and different intersections interact only through the network-level model.

\subsection{Hierarchical Action-Conditioned Dynamics}
\label{subsec:hier_transition}

The model $f_\theta$ consists of two sequentially coupled predictors. The \emph{network-level model} operates on $\mathcal{G}^{\mathrm{net}}$ and predicts the dynamic variables of each intersection; the \emph{lane-level model} operates on the graphs $\mathcal{G}^{\mathrm{lane}}_v$ and predicts those of the controlled lanes, including the quantities of the stage cost~\eqref{eq:stage_cost}. With $X^{\mathrm{net}}(t)$ the network-level node and edge features and $X^{\mathrm{lane}}(t)$ the lane-level features,
\begin{align}
\hat{Y}^{\mathrm{net}}(t+1)
&= f^{\mathrm{net}}_{\theta_{\mathrm{net}}}
\big(\mathcal{G}^{\mathrm{net}},X^{\mathrm{net}}(t),u(t)\big),
\label{eq:net_stage}\\
\hat{Y}^{\mathrm{lane}}(t+1)
&= f^{\mathrm{lane}}_{\theta_{\mathrm{lane}}}
\big(\{\mathcal{G}^{\mathrm{lane}}_v\},
X^{\mathrm{lane}}(t),\notag\\
&\hspace{3.1em}\hat{Y}^{\mathrm{net}}(t+1),u(t)\big),
\label{eq:int_stage}
\end{align}
where $\hat{Y}^{\mathrm{net}}$ and $\hat{Y}^{\mathrm{lane}}$ collect the predicted network- and lane-level variables. At each prediction step, the lane variables are aggregated into network-edge features, the network-level model predicts the intersection states under the candidate joint phase, and these predictions provide context for the lane-level model; the outputs update the information state for the next rollout step.

Signal actions condition the model at three points, which distinguish the discharge of an intersection from its arrivals. The candidate local phase $u_v(t)$ enters both predictions at intersection $v$ and determines which of its lanes discharge. The candidate phase $u_{v_i}(t)$ of an upstream intersection conditions the message along each network edge $(v_i,v_j)$, so that the prediction at $v_j$ depends on what $v_i$ releases toward it. Since the lane-level graphs do not connect intersections, the lane-level model receives the arrival information directly, as the previous phase $u_\ell^{\mathrm{up}}(t-1)$ of the nearest upstream signalized intersection feeding lane $\ell$, taken from the planned actions in later rollout steps. The model thus predicts the evolution of traffic under alternative signal decisions rather than a single forecast. This formulation is motivated by the interactions represented in store-and-forward models~\cite{gazis1963oversaturated,aboudolas2009store}, in which queue evolution depends on upstream arrivals, signal-dependent discharge, turning movements and storage; the learned model estimates their combined effects from traffic transitions instead of specifying them.

\subsection{Inputs and Predicted Quantities}
\label{subsec:feature_design}
\label{subsec:targets}

The inputs combine lagged measurements over $H$ steps with static attributes, incident descriptors and known signal information (Supplementary Material, Sec.~S1-A). Network-level node features contain the intersection pressure, average delay, incoming vehicle count and lane-occupancy statistics, and each network edge $(v_i,v_j)$ aggregates the current variables of the controlled lanes at $v_j$ fed by $v_i$. Lane-level features contain the queue, accumulated waiting time, incoming vehicle count and speed, and occupancy, the number of vehicles that entered the windows of the nearest upstream signalized intersection during the last interval, together with the upstream halt of the lane's approach where a zone exists, the network context $\hat{Y}^{\mathrm{net}}_{v(\ell)}(t+1)$ predicted for the lane's intersection, and the embeddings of the two phases. Incident descriptors include blockage indicators, the blockage position and the elapsed time since onset and clearance; when a blockage shortens the usable part of a lane window, occupancy is normalized by the remaining effective length. The targets are the dynamic variables at the next step. Although the upstream halt is measured per approach, the lane-level model predicts it for each controlled lane, and the predictions are averaged to the approach-level quantity in the objective:
\begin{equation}
\label{eq:upstream_avg}
\hat{q}^{\mathrm{up}}_a(t+1)
=
\frac{1}{|\mathcal{L}_a|}
\sum_{\ell\in\mathcal{L}_a}
\hat{q}^{\mathrm{up}}_\ell(t+1).
\end{equation}

\subsection{Model Architecture}
\label{subsec:net_model}
\label{subsec:int_model}

Both predictors encode the lagged dynamic variables of each node with an LSTM, concatenate the encoding with static attributes, incident descriptors and phase embeddings, and apply three GATv2 layers~\cite{brody2021attentive} with residual connections (Supplementary Material, Sec.~S1, Fig.~S1). In the network-level model, the edge features are encoded together with the candidate phase of the upstream intersection, so that the attention weights can vary with traffic, incident and signal conditions, and the prediction head is a mixture of two experts combined by an input-dependent sigmoid gate, trained through the prediction loss without regime labels. The lane-level model operates on $\mathcal{G}^{\mathrm{lane}}_v$, with a reserved embedding for lanes without an upstream signal, and an MLP head predicts the next-step lane variables. Dimensions and hyperparameters are given in Supplementary Material, Sec.~S3.

\subsection{Training and Multi-Step Prediction}
\label{subsec:training}

The model is trained offline on recorded one-step transitions from simulated episodes with incidents operated by MaxPressure; part of the episodes can add random phase exploration, which exposes the model to decisions that MaxPressure rarely selects (the data used for each network are specified in Section~\ref{subsec:config}). Training is sequential: the network-level model is trained with a Huber loss and frozen, and the lane-level model is then trained with a squared-error loss on the predictions of the frozen network-level model, which exposes it to the estimated context available during inference (Supplementary Material, Sec.~S1-E). During planning, the model is applied recursively according to~\eqref{eq:dynamics}: the predicted variables are fed back into the lag windows, the previous upstream phase is taken from the observed signal history at the first step and from the candidate sequence thereafter, the upstream incoming count is replaced by the incoming count predicted for the upstream intersection, incident information is held, and elapsed-time descriptors advance. The model is not adapted online: it supports replanning after detected disruptions, but not the identification of previously unseen dynamics.

\section{Coordinated Receding-Horizon Planning}
\label{sec:planning}

This section presents the approximate solution of~\eqref{eq:mpc_obj}--\eqref{eq:control_hold} with the learned model $f_\theta$ as the predictor $\hat{F}$. A candidate joint phase sequence is evaluated by rolling out the learned model from $X(t)$ and accumulating the stage cost~\eqref{eq:stage_cost} over the prediction horizon; independent rollouts are evaluated in batches. The rollout is deterministic for a given information state and control sequence, apart from numerical variations in GPU execution, which are discussed after Proposition~\ref{prop:monotone_descent}.

\subsection{Decomposition by Intersection}
\label{subsec:distributed_mpc}

The joint decision space $\mathcal{U}=\prod_{v\in\mathcal{V}}\mathcal{U}_v^{n_u}$ grows exponentially with the number of intersections and the control horizon, so the decision variables are decomposed by intersection. Each intersection $v$ acts as an agent choosing its phase sequence $U_v=(u_{v,0|t},\ldots,u_{v,n_u-1|t})\in\mathcal{U}_v^{n_u}$, and the joint plan is written $U=(U_v,U_{-v})$, where $U_{-v}$ collects the sequences of all other intersections. Every candidate is evaluated with the shared cost $J_t(U_v,U_{-v})$, whose rollout predicts the effects of an intersection's decisions across the whole network; the decomposition thus concerns the optimization variables, not the dynamics or the objective. The coordination problem is the finite game $\Gamma_t=\langle\mathcal{V},\{\mathcal{U}_v^{n_u}\}_{v\in\mathcal{V}},J_t\rangle$, in which all agents minimize the same cost; it is an identical-interest game and therefore an exact potential game with potential $J_t$~\cite{monderer1996potential}. The best-response set of intersection $v$ is
\begin{equation}
\label{eq:best_response}
    \mathcal{B}_v(U_{-v})=\arg\min_{U_v\in\mathcal{U}_v^{n_u}}J_t(U_v,U_{-v}).
\end{equation}

\begin{definition}[$\varepsilon$-equilibrium]
\label{def:equilibrium}
Let $\varepsilon\geq0$ and $\mathcal{S}_v\subseteq\mathcal{U}_v^{n_u}$ be a candidate set for each intersection $v\in\mathcal{V}$. A joint plan $U^\star$ is an $\varepsilon$-equilibrium of $\Gamma_t$ relative to $\{\mathcal{S}_v\}$ if
\begin{equation}
\label{eq:eps_equilibrium}
    J_t(U_v,U^\star_{-v})\geq J_t(U^\star)-\varepsilon,
    \quad \forall U_v\in\mathcal{S}_v,\;\forall v\in\mathcal{V}.
\end{equation}
For $\varepsilon=0$ and $\mathcal{S}_v=\mathcal{U}_v^{n_u}$ for every $v$, this reduces to a pure Nash equilibrium.
\end{definition}

Every global minimizer of the shared cost is a Nash equilibrium, but not conversely: at an equilibrium, no single intersection can improve the objective unilaterally, although coordinated changes by several intersections may.

\subsection{Iterated Best Response with Acceptance Test}
\label{subsec:ibr}

Starting from an admissible joint plan $U^{(0)}$, random or from a heuristic controller (Section~\ref{subsec:config}), the intersections are visited in a fixed order over successive passes, each optimizing against the most recent sequences of the others (Gauss--Seidel). Let $k$ index the individual updates and $v$ denote the intersection updated at step $k$. Since exact minimization over $\mathcal{U}_v^{n_u}$ is combinatorial, the best response is approximated over a finite candidate set $\mathcal{S}_v^{(k)}\subseteq\mathcal{U}_v^{n_u}$ generated by probabilistic beam search (Section~\ref{subsec:optimization}):
\begin{equation}
\label{eq:approx_br}
    \tilde{U}_v^{(k)}\in\arg\min_{U_v\in\mathcal{S}_v^{(k)}}J_t(U_v,U_{-v}^{(k)}).
\end{equation}
With $\tilde{U}^{(k)}=(\tilde{U}_v^{(k)},U_{-v}^{(k)})$, the update is accepted only if it reduces the shared full-horizon cost by more than a threshold $\varepsilon\geq0$:
\begin{equation}
\label{eq:acceptance}
    U^{(k+1)}=
    \begin{cases}
        \tilde{U}^{(k)}, & J_t(\tilde{U}^{(k)})<J_t(U^{(k)})-\varepsilon,\\
        U^{(k)}, & \text{otherwise}.
    \end{cases}
\end{equation}

\begin{proposition}[Monotone descent and finite termination]
\label{prop:monotone_descent}
Suppose that, at a fixed control step $t$, the predicted cost $J_t:\mathcal{U}\rightarrow\mathbb{R}$ is a fixed deterministic function of the joint plan. For any realization of the candidate sets generated by the search, the updates~\eqref{eq:approx_br}--\eqref{eq:acceptance} satisfy:

\emph{(i) Monotone descent:} $J_t(U^{(k+1)})\leq J_t(U^{(k)})$ for every update $k$, and every accepted update decreases the cost by more than $\varepsilon$.

\emph{(ii) Finite termination:} The number of accepted updates is finite. For $\varepsilon>0$, it is bounded above by $(J_t(U^{(0)})-\min_{U\in\mathcal{U}}J_t(U))/\varepsilon$. For $\varepsilon=0$, there are at most $|\mathcal{U}|-1$ accepted updates. Consequently, if the procedure continues without an external stopping limit, a complete pass without an accepted update occurs after finitely many passes.

\emph{(iii) Candidate-set equilibrium:} If a complete pass ends without an accepted update, the resulting plan $U^\star$ is an $\varepsilon$-equilibrium relative to the candidate sets evaluated during that pass. If $\varepsilon=0$ and every best-response search is exhaustive, $U^\star$ is a Nash equilibrium.
\end{proposition}

The proof is given in the Appendix. Any returned plan satisfies $J_t(U^\star)\leq J_t(U^{(0)})$, so a plan initialized from a heuristic controller cannot worsen its predicted cost. The guarantee concerns the predicted objective at one control step: it does not establish global optimality, closed-loop stability or improvement in actual traffic conditions, and the equilibrium property applies only to the candidate sets of a pass without accepted updates. In implementation, the procedure stops after such a pass, when the relative improvement over a pass falls below a tolerance $\delta$, or after $K_{\max}$ passes; the latter two conditions bound computation without establishing equilibrium, whereas the non-increasing cost holds under all three. Since parallel GPU execution can make repeated evaluations differ slightly, the implementation compares and stores the evaluated costs of accepted plans, which are therefore non-increasing by construction. The scheme is related to cooperative distributed MPC, in which local controllers optimize a shared objective~\cite{stewart2010cooperative}; instead of convex combinations of continuous inputs, the discrete phase decisions are coordinated by sequential best responses with an explicit acceptance test, which preserves monotone descent without a convex decision space. Every sequence in $\mathcal{U}$ is admissible, since the yellow interval is inserted within the control interval and no minimum green time is imposed, so the problem is feasible at every step.

\subsection{Best-Response Computation}
\label{subsec:optimization}

Algorithm~\ref{alg:ibr_pbs} summarizes the planning procedure at one control step. Each best response is computed by probabilistic beam search (Algorithm~S1 of the Supplementary Material), which constructs the sequence of intersection $v$ step by step with $U_{-v}$ fixed. At each depth, the extensions of the $W$ retained partial sequences are scored by partial rollouts, $N_s$ of them are sampled from a temperature-controlled distribution over their partial costs mixed with a uniform distribution, and the $W$ sampled extensions with the lowest partial cost are retained. Sampling keeps sequences whose partial cost is higher but whose full-horizon cost may be lower once queues discharge. The final beam, extended to the prediction horizon by~\eqref{eq:control_hold}, forms the candidate set $\mathcal{S}_v^{(k)}$ and is evaluated with full-horizon rollouts, so that the search considers at most $W|\mathcal{U}_v|$ extensions per depth instead of $|\mathcal{U}_v|^{n_u}$ complete sequences. Each coordination pass performs $|\mathcal{V}|$ sequential searches with a number of model evaluations that does not depend on the network size, but each evaluation is a network-wide prediction whose cost grows linearly with the number of intersections, so that the work per pass grows quadratically; batched GPU evaluation can make the measured time grow more slowly (Section~\ref{subsec:computation}).

\begin{algorithm}[t]
\caption{Coordinated receding-horizon planning with probabilistic beam search (one control step)}
\label{alg:ibr_pbs}
\small
\begin{algorithmic}[1]
\Require Information state $X(t)$, phase sets $\{\mathcal{U}_v\}_{v\in\mathcal{V}}$, horizons $n_u\leq n_y$, initial plan $U^{(0)}$, acceptance threshold $\varepsilon\geq0$, tolerance $\delta$, constant $\eta>0$, maximum passes $K_{\max}$
\Ensure Joint plan $U^\star$
\State $U\gets U^{(0)}$; \quad $J_{\mathrm{cur}}\gets J_t(U)$ \Comment{full-horizon rollout} \label{alg1:init}
\For{$p=1,\ldots,K_{\max}$} \label{alg1:pass}
    \State $J_{\mathrm{prev}}\gets J_{\mathrm{cur}}$; \quad $\texttt{accepted}\gets\texttt{false}$
    \For{each $v\in\mathcal{V}$ in a fixed order} \label{alg1:sweep}
        \State $(\tilde{U}_v,\tilde{J})\gets\textsc{BeamSearch}(v,U,X(t))$ \Comment{Algorithm~S1} \label{alg1:br}
        \If{$\tilde{J}<J_{\mathrm{cur}}-\varepsilon$} \label{alg1:accept}
            \State $U_v\gets\tilde{U}_v$; \quad $J_{\mathrm{cur}}\gets\tilde{J}$; \quad $\texttt{accepted}\gets\texttt{true}$ \label{alg1:update}
        \EndIf
    \EndFor
    \If{$\texttt{accepted}=\texttt{false}$}
        \State \textbf{break} \Comment{no accepted update} \label{alg1:fixed}
    \EndIf
    \If{$(J_{\mathrm{prev}}-J_{\mathrm{cur}})/\max(|J_{\mathrm{prev}}|,\eta)<\delta$}
        \State \textbf{break} \Comment{relative improvement tolerance} \label{alg1:tol}
    \EndIf
\EndFor
\State \Return $U^\star\gets U$ \label{alg1:return}
\end{algorithmic}
\end{algorithm}

\section{Experimental Design}
\label{sec4:experiment}
\label{subsec:implementation}

\subsection{Networks and Observation Coverage}
\label{subsec:simulation}

The experiments use two networks with contrasting controlled-lane geometry (Table~\ref{tab:networks}): the $4\times4$ grid of the RESCO benchmark~\cite{ault2021reinforcement} and the Ingolstadt network of the InTAS benchmark~\cite{lobo2020intas}, whose layouts are shown in Supplementary Material, Sec.~S2. The controlled lanes of the grid exceed the measurement range of 200~m, whereas 154 of the 158 controlled lanes on Ingolstadt are shorter. Upstream zones cover 56 of the 67 Ingolstadt approaches (Supplementary Material, Sec.~S1-G) and hold 57--62\% of the queued length, depending on the controller (Supplementary Material, Sec.~S7-C). The networks also differ in topology and demand, so their comparison does not isolate lane geometry. The experiments run in T-REX~\cite{nguyen2025robustness} on SUMO~\cite{lopez2018sumo}, with 1~s simulation steps, a phase decision every 10~s and a 3~s yellow interval between different phases. An incident is a stopped vehicle that blocks one or more lanes, around which informed drivers reroute (Supplementary Material, Sec.~S2).

\begin{table}[t]
\centering
\caption{Networks and evaluation scenarios.}
\label{tab:networks}
\small
\setlength{\tabcolsep}{2pt}
\begin{tabular}{lcc}
\toprule
 & \textbf{Grid $4 \times 4$} & \textbf{Ingolstadt} \\
\midrule
Signalized intersections & 16 & 21 \\
Approaches / controlled lanes & 64 / 192 & 67 / 158 \\
Lanes per approach & 3 & 1--4 \\
Controlled lane length (m) & 273--286 & 0.2--271 (med.~24) \\
Green phases per signal & 8 & 2--4 \\
Network-level graph edges & 48 & 75 \\
Vehicles per episode & 1,473--1,556 & 4,240 \\
Episode duration (s) & 3,600$^{\dagger}$ & 3,560 \\
\midrule
$L^{\mathrm{meas}}$ / $R$ (m) & 200 / -- & 200 / 300 \\
\midrule
Random-incident scenarios & E1--E5 & S1--S5 \\
\quad incidents per scenario & 2 & 2 \\
Predefined-incident scenarios & EX2A--D (9) & -- \\
\quad incidents per scenario & 1 (0 in EX2D-0) & -- \\
Trials per scenario & 5 & 5 \\
\bottomrule
\multicolumn{3}{l}{\footnotesize $^{\dagger}$Including a 100~s warm-up under the fixed signal program.}
\end{tabular}
\end{table}

\subsection{Incident Scenarios}
\label{subsec:scenarios}

Evaluation scenarios are not contained in any training data, and the incidents of a scenario are identical in all its trials and for all controllers. Each network has five random-incident scenarios with two incidents each (E1--E5 and S1--S5), the most severe of which, E3, includes a full closure that outlasts the episode; on the grid, eight further scenarios each vary the location, duration or number of blocked lanes of a single incident, and a ninth, EX2D-0, is an incident-free reference. Trials differ in the simulation seed and, in the predefined scenarios, in the demand pattern; the scenario definitions are given in Supplementary Material, Sec.~S2. Incident information is available as stated in Assumption~\ref{ass:incident}; detection delays and unknown incident locations are not evaluated.

\subsection{Controllers and Comparative Design}
\label{subsec:baselines}

Table~\ref{tab:baselines} summarizes the controllers, which all choose among the same green phases every 10~s. Fixed-Time control serves the green phases in order for 30~s each, and MaxPressure~\cite{varaiya2013max} selects the phase with the largest pressure in the lane windows. Five reinforcement-learning (RL) controllers, implemented in the RESCO framework~\cite{ault2021reinforcement}, span the coordination mechanisms of the multi-agent literature: IDQN and IPPO~\cite{Ault2020Learning}, MPLight~\cite{chen2020toward}, FMA2C~\cite{ma2020feudal} and CoSLight~\cite{ruan2024coslight}. Each was trained with and without random incidents, and and the condition with the lower mean average travel time (ATT; Section~\ref{subsec:evaluation}) on the evaluation runs is reported, which gives the RL baselines an optimistic estimate; the traveler and the network ATT select the same condition for every method on both networks (Supplementary Material, Sec.~S4). The RL controllers and MaxPressure observe the lane windows only and receive no incident information, whereas GNN-MPC also uses the incident information of Assumption~\ref{ass:incident} and, on Ingolstadt, the upstream zones (Table~\ref{tab:baselines}). Withholding the incident descriptors from GNN-MPC (variant B) raises its traveler ATT by 2.1~s on the grid and has no resolved effect on Ingolstadt (Section~\ref{subsec:ablation}), much less than its margins over these baselines; the role of the upstream zones is examined by the observation-related comparisons below.

Three types of comparison on Ingolstadt address the roles of upstream observation and learned prediction. The \emph{observation-related comparisons} contrast MaxPressure with MaxPressure-U, which adds the halting count of the upstream zone to the incoming pressure term, and GNN-MPC with GNN-MPC (No Upstream), whose lane-level model is retrained without the upstream-zone input and target and whose objective omits the upstream term, while the inputs from the windows of neighboring intersections are retained; they compare configurations rather than forming a factorial design. The \emph{information-matched comparison} contrasts MaxPressure-U with GNN-MPC, which have comparable observation coverage but also differ in optimization procedure, objective and explicit incident information; variant B, which withholds the incident descriptors from GNN-MPC, is a sensitivity check for this asymmetry. The \emph{analytical-versus-learned comparison} replaces the learned model with a calibrated store-and-forward predictor~\cite{aboudolas2009store,lin2011fast} under the same traffic measurements, objective, horizons and planner (SF-MPC), and with an extension that adds arrivals into the upstream zone, storage limits on the actual window lengths and spillback (SF-MPC-Ext); both are calibrated on the training episodes only (Supplementary Material, Sec.~S9).

\begin{table}[t]
\centering
\caption{Information, prediction and decision procedure of the controllers. W: lane windows; W+U: lane windows and upstream zones, which exist on Ingolstadt only, so that W+U reduces to W on the grid; IBR: iterated best response (Section~\ref{sec:planning}). MaxPressure-U, GNN-MPC (No Upstream) and SF-MPC-Ext are evaluated on Ingolstadt only.}
\label{tab:baselines}
\scriptsize
\setlength{\tabcolsep}{2pt}
\begin{tabular}{@{}lccll@{}}
\toprule
\textbf{Controller} & \textbf{Obs.} & \textbf{Incid.} & \textbf{Prediction} & \textbf{Decision procedure} \\
\midrule
Fixed-Time & -- & no & -- & fixed phase order \\
MaxPressure & W & no & -- & max.\ pressure \\
RL (5 methods) & W & no & -- & trained policy \\
MaxPressure-U & W+U & no & -- & max.\ pressure + upstream \\
\midrule
GNN-MPC (No Upstream) & W & yes & learned, no zone & IBR, no zone term \\
SF-MPC & W+U & yes & store-and-forward & IBR, cost~\eqref{eq:stage_cost} \\
SF-MPC-Ext & W+U & yes & extended SF & IBR, cost~\eqref{eq:stage_cost} \\
GNN-MPC & W+U & yes & learned two-level & IBR, cost~\eqref{eq:stage_cost} \\
\bottomrule
\end{tabular}
\end{table}

\subsection{GNN-MPC Configuration}
\label{subsec:config}

The learned model is trained per network on episodes with random incidents operated by MaxPressure: 100 grid and 296 Ingolstadt episodes, split by episode into training, validation and test sets, with random phase exploration in one third of the Ingolstadt training episodes. The control and prediction horizons, set in advance, are 20~s on the grid and 30 and 60~s on Ingolstadt; they were chosen from the network structure and preliminary trials on the training episodes to cover the typical travel time between neighboring signalized intersections, so that the effect of a phase decision reaches the neighboring intersections within the prediction horizon, while keeping the planning time close to the control interval. Longer and shorter horizons are tested as variants E and H (Section~\ref{subsec:ablation}). On the grid, the stage cost is the total predicted lane queue; on Ingolstadt, it also includes the upstream halt and the waiting-time term ($\lambda=1$, with scales equal to the standard deviations of the targets). The initial plan is random on the grid and, on Ingolstadt, given by the MaxPressure rule applied at each step to the predicted states, so that by Proposition~\ref{prop:monotone_descent} the selected plan has a predicted cost no higher than that plan. Variants are retrained only as far as their change requires (Supplementary Material, Sec.~S3). All evaluation runs use fixed seeds and deterministic GPU execution.

\subsection{Metrics and Statistical Analysis}
\label{subsec:evaluation}

Let $\mathcal{S}$ be the vehicles scheduled to depart within the evaluation window, excluding the incident vehicles. Each is either \emph{admitted}, i.e., inserted into the network, or \emph{never inserted}. On Ingolstadt, the simulation runs 43~s beyond the window, and an admitted vehicle either completes its trip within the window, completes it during the extension, or is still in the network at the end of the simulation; on the grid, the simulation ends with the window. Only trips completed within the window count in the \emph{throughput}, and all times are censored at the end of the window; teleported vehicles remain in these counts with their full travel time. Since censoring only shortens the times of unfinished trips, it tends to favor controllers that leave more of the demand unserved; excluding the vehicles scheduled near the end of the window leaves the sign of every paired comparison unchanged and widens each difference by at most 7\% (Supplementary Material, Sec.~S7-F). This accounting separates \emph{admission}, i.e., how much of the scheduled demand enters the network, from \emph{service}, i.e., how much of the admitted demand completes its trip.

The \emph{traveler ATT} is the mean travel time over all of $\mathcal{S}$ from the scheduled departure, so that it includes the delay before insertion and charges a never-inserted vehicle the remaining window. It is the primary travel-time metric, and the results are interpreted together with the throughput, the completed trips and the never-inserted vehicles. The \emph{network ATT}, the mean travel time of the admitted vehicles from their insertion, counts only the demand that a controller admits: under heavy demand, when departures are delayed at the network boundary, a controller that admits less of the scheduled demand can obtain a lower network ATT without serving that demand better. The network ATT and the corresponding waiting time are therefore reported for reference in the Supplementary Material, Secs.~S6 and~S7, together with further metrics. On the grid, every vehicle is admitted, and the two ATT differ by at most 0.2~s.

Controllers are compared by paired differences, with pairs sharing the scenario, seed, demand and incidents, reported as the mean with its 95\% Student-$t$ interval and the number of pairs in which the first controller is better. Since the runs of a scenario share its incidents, the principal differences are also given with the scenario as the unit of analysis (four degrees of freedom); a difference is called robust under clustering when this interval excludes zero. All intervals are conditional on the evaluated scenarios. The principal comparisons use 25 runs per network (five trials of each of E1--E5 and S1--S5); on Ingolstadt, the four observation-related configurations use the same 25 runs. The analytical-versus-learned comparisons and the component variants use nine Ingolstadt runs (S2, S3, S5) and 15 grid runs (E1, E3, E5, EX2C-2, EX2D-0), with three trials each. Each learning-based method is represented by one trained model, so the reported variability reflects the evaluation seeds rather than repeated training. Per-scenario results and interactive visualizations of the experiments on both networks, including the individual runs, are available online.\footnote{\begin{tabular}[t]{@{}l@{}}Results page: \url{https://ai-prolight.eu/news/gnn_mpc/}\\ Experiment dashboard: \url{https://ai-prolight.eu/dashboard.html}\end{tabular}}

\section{Results}
\label{sec:results}

\subsection{Overall Closed-Loop Performance}
\label{subsec:results_overall}
\label{subsec:results_grid}
\label{subsec:results_ingolstadt}

On the grid, GNN-MPC achieves the lowest traveler ATT in every scenario and in all 25 trial pairs against every baseline, and the highest mean throughput (Table~\ref{tab:grid_main}): 14--16\% below MaxPressure under the mild incidents, 26\% below it under the severe incident of E3, and 6--13\% below the strongest learning-based baseline of each scenario. The advantage is present without incidents and grows relative to MaxPressure under them: in incident-free runs with the same demand and seeds, the traveler ATT of GNN-MPC is already 22.4~s lower than that of MaxPressure, and the incidents add 15.3~s to it against 27.0~s for MaxPressure, although Fixed-Time and FMA2C lose slightly less under the severe incident, from much higher levels (Supplementary Material, Sec.~S6-B). For the vehicles whose routes expose them to an incident, the margin over MaxPressure (26--93~s) exceeds the network-wide margin in every scenario. The predefined incidents confirm the ranking across incident locations, durations and numbers of blocked lanes (Supplementary Material, Sec.~S6), except that under a prolonged two-lane partial blockage the margin over MaxPressure and MPLight narrows.

\begin{table}[t]
\centering
\caption{Grid, E1--E5: traveler ATT, mean over five trials, and throughput, i.e., trips completed per run out of 1,473 scheduled vehicles, mean over E1--E5. $\Delta$: paired difference GNN-MPC minus method in traveler ATT over the 25 trial pairs, mean $\pm$ half-width of the unclustered 95\% Student-$t$ interval; GNN-MPC is better in all 25 pairs for every baseline. NI/I: RL training without/with incidents.}
\label{tab:grid_main}
\scriptsize
\setlength{\tabcolsep}{1.6pt}
\begin{tabular}{lrrrrrrr}
\toprule
 & \multicolumn{5}{c}{\textbf{Traveler ATT (s)}} & & \\
\cmidrule(lr){2-6}
\textbf{Method} & \textbf{E1} & \textbf{E2} & \textbf{E3} & \textbf{E4} & \textbf{E5} & \textbf{Thr.\ (veh)} & $\boldsymbol{\Delta}$ \textbf{(s)} \\
\midrule
Fixed-Time & 292.0 & 291.9 & 348.9 & 291.5 & 295.2 & 1,412 & $-149.8 \pm 2.1$ \\
MaxPressure & 162.1 & 163.1 & 282.1 & 163.4 & 170.4 & 1,428 & $-34.2 \pm 8.5$ \\
IDQN (NI) & 152.4 & 219.5 & 572.5 & 149.0 & 155.5 & 1,365 & $-95.7 \pm 58.3$ \\
MPLight (I) & 160.7 & 162.2 & 233.4 & 160.6 & 166.2 & 1,440 & $-22.6 \pm 0.7$ \\
IPPO (NI) & 160.2 & 167.9 & 341.2 & 158.7 & 167.8 & 1,424 & $-45.1 \pm 18.7$ \\
FMA2C (NI) & 213.8 & 215.2 & 276.5 & 214.9 & 220.4 & 1,427 & $-74.1 \pm 1.2$ \\
CoSLight (NI) & 190.4 & 191.3 & 265.4 & 190.3 & 195.2 & 1,430 & $-52.5 \pm 1.3$ \\
GNN-MPC & \textbf{139.4} & \textbf{140.4} & \textbf{207.8} & \textbf{139.4} & \textbf{143.3} & \textbf{1,448} & -- \\
\bottomrule
\end{tabular}
\end{table}

On Ingolstadt, GNN-MPC also has the lowest traveler ATT, the highest throughput and the fewest never-inserted vehicles (Table~\ref{tab:ingol_main}). Its traveler ATT is 75--244~s lower than that of every baseline, in all 25 pairs, including 76.7~s below MaxPressure, and it completes 102 more trips per run than the next-best controller, FMA2C. Part of the difference lies in admission: MaxPressure and MPLight leave 193 and 162 vehicles per run outside the network, against 12 for GNN-MPC.

\begin{table}[t]
\centering
\caption{Ingolstadt, S1--S5: means per run over 25 runs. Thr.: trips completed within the window, out of 4,240 scheduled vehicles; Never ins.: vehicles never inserted. $\Delta$: paired difference GNN-MPC minus method in traveler ATT, mean $\pm$ half-width of the unclustered 95\% Student-$t$ interval; GNN-MPC is better in all 25 pairs for every baseline. NI/I: RL training without/with incidents. Further metrics in Supplementary Material, Sec.~S7.}
\label{tab:ingol_main}
\scriptsize
\setlength{\tabcolsep}{3pt}
\begin{tabular}{lrrrr}
\toprule
\textbf{Method} & \textbf{Thr.\ (veh)} & \textbf{Never ins.\ (veh)} & \textbf{Trav.\ ATT (s)} & $\boldsymbol{\Delta}$ \textbf{(s)} \\
\midrule
Fixed-Time & 3,808 & 27 & 332.8 & $-100.1 \pm 7.3$ \\
MaxPressure & 3,753 & 193 & 309.5 & $-76.7 \pm 7.4$ \\
IDQN (I) & 3,802 & 20 & 307.9 & $-75.1 \pm 8.3$ \\
MPLight (NI) & 3,701 & 162 & 339.5 & $-106.7 \pm 9.9$ \\
IPPO (I) & 3,799 & 25 & 322.7 & $-89.9 \pm 6.2$ \\
FMA2C (I) & 3,856 & 29 & 313.4 & $-80.6 \pm 9.2$ \\
CoSLight (I) & 3,432 & 44 & 477.1 & $-244.3 \pm 18.1$ \\
GNN-MPC & \textbf{3,958} & \textbf{12} & \textbf{232.8} & -- \\
\bottomrule
\end{tabular}
\end{table}

The advantage is not uniform. Relative to MaxPressure, it arises at entries with short controlled lanes, whose queues extend into the upstream zones: in a decomposition of the traveler delay by entry approach, these entries account for the whole difference (Supplementary Material, Sec.~S7-E). In time, the advantage over MaxPressure-U, which observes the same zones, is concentrated in the incident windows ($-44.4$~s [$-69.1$, $-19.7$], clustered by scenario), whereas the difference in the lightly loaded period before the incidents is not resolved (Supplementary Material, Sec.~S7-D). These patterns associate the advantage with demand upstream of short lanes and with high load, but they do not separate the observation from the controller that uses it, which Section~\ref{subsec:observation} examines.

\subsection{Upstream Observation, Admission and Service}
\label{subsec:observation}

Adding the upstream zones to pressure-based control changes admission more than service (Table~\ref{tab:observation}). MaxPressure-U admits 178 more vehicles per run than MaxPressure but completes only 87 more trips, with 94 more vehicles still in the network at the end; its traveler ATT falls by 41.3~s. In the predictive family, the upstream representation changes admission by a similar amount, with different consequences. GNN-MPC (No Upstream) holds about as much demand at the boundary as MaxPressure, with a traveler ATT not distinguishable from it, whereas GNN-MPC admits 161 more vehicles and completes 163 more trips without a resolved increase in the vehicles left in the network, which lowers the traveler ATT by 73.4~s. The difference between the two upstream effects is resolved under clustering for the traveler ATT (32~s) and the vehicles left in the network, and only at the run level for the completed trips. Since the upstream information enters the two families differently, these are comparisons of configurations rather than a factorial design; nevertheless, neither MaxPressure-U nor GNN-MPC (No Upstream) reproduces the admission and service of the complete system.

\begin{table*}[t]
\centering
\caption{Ingolstadt, S1--S5 (25 runs): upstream observation, admission and service. (a)~Means per run; the four vehicle categories are mutually exclusive and sum to the 4,240 scheduled vehicles. (b)~Paired differences, mean with the 95\% interval clustered by scenario. Last row: (GNN-MPC $-$ No Upstream) $-$ (MaxPr.-U $-$ MaxPressure), per run. No Upstream: GNN-MPC (No Upstream); MaxPr.-U: MaxPressure-U. Unclustered intervals, wins, and the travel and waiting times of the admitted vehicles in Supplementary Material, Sec.~S8, which also links the per-scenario values.}
\label{tab:observation}
\scriptsize
\setlength{\tabcolsep}{2pt}
\begin{tabular}{lrrrrr}
\toprule
 & \multicolumn{2}{c}{\textbf{Completed}} & \textbf{Still in network} & \textbf{Never} & \textbf{Traveler} \\
\cmidrule(lr){2-3}
 & \textbf{within window} & \textbf{in extension} & \textbf{at simulation end} & \textbf{inserted} & \textbf{ATT (s)} \\
\midrule
\multicolumn{6}{l}{\emph{(a) Means per run}} \\
MaxPressure & 3753.0 & 52.3 & 241.8 & 193.0 & 309.5 \\
MaxPressure-U & 3839.7 & 50.3 & 335.3 & 14.7 & 268.2 \\
GNN-MPC (No Upstream) & 3794.4 & 45.2 & 226.8 & 173.6 & 306.1 \\
GNN-MPC & 3957.8 & 50.6 & 219.2 & 12.4 & 232.8 \\
\midrule
\multicolumn{6}{l}{\emph{(b) Paired differences}} \\
MaxPr.-U $-$ MaxPressure & 86.7 [0.7, 172.7] & $-$2.0 [$-$8.3, 4.4] & 93.5 [14.8, 172.2] & $-$178.2 [$-$196.4, $-$160.1] & $-$41.3 [$-$60.5, $-$22.1] \\
No Upstream $-$ MaxPressure & 41.4 [$-$25.3, 108.1] & $-$7.1 [$-$15.8, 1.5] & $-$14.9 [$-$47.0, 17.2] & $-$19.4 [$-$48.5, 9.7] & $-$3.4 [$-$17.5, 10.8] \\
GNN-MPC $-$ No Upstream & 163.4 [128.2, 198.6] & 5.4 [2.0, 8.9] & $-$7.7 [$-$29.0, 13.6] & $-$161.2 [$-$182.5, $-$139.8] & $-$73.4 [$-$83.7, $-$63.0] \\
GNN-MPC $-$ MaxPr.-U & 118.1 [55.9, 180.4] & 0.3 [$-$4.5, 5.1] & $-$116.1 [$-$182.5, $-$49.7] & $-$2.3 [$-$5.2, 0.6] & $-$35.5 [$-$59.9, $-$11.0] \\
Diff.\ of upstream effects & 76.7 [$-$31.3, 184.7] & 7.4 [$-$1.2, 16.0] & $-$101.2 [$-$184.6, $-$17.8] & 17.1 [$-$11.3, 45.5] & $-$32.1 [$-$59.8, $-$4.3] \\
\bottomrule
\end{tabular}
\end{table*}

Under comparable observation coverage and nearly identical admission, GNN-MPC completes 118 more trips than MaxPressure-U and leaves 116 fewer in the network, both robust under clustering, and its traveler ATT is 35.5~s (13\%) lower [$-59.9$, $-11.0$]. Differences in the number of admitted vehicles are thus unlikely to explain this service advantage, although the admitted populations need not be identical. Unlike the margin over MaxPressure, the advantage is spread over many approaches, the largest contributing about a third (Supplementary Material, Sec.~S7-E). The comparison credits the complete predictive controller rather than the learned dynamics alone, since the two controllers also differ in objective, horizon, optimization procedure and explicit incident information; withholding the incident descriptors (variant B, Section~\ref{subsec:ablation}) gives no resolved difference in traveler ATT ($-0.2$~s [$-8.3$, $7.9$]), an interval compatible with effects in either direction. Section~\ref{subsec:sf} therefore replaces the predictive model while retaining the planner and objective.

\subsection{Learned versus Analytical Prediction}
\label{subsec:sf}
\label{subsec:surrogate_accuracy}

With the same traffic measurements, objective and planner, replacing the learned dynamics changes the closed-loop outcome substantially on Ingolstadt. This substitution is evaluated separately from the comparisons of Section~\ref{subsec:results_overall}, on the nine Ingolstadt runs (S2, S3 and S5, selected by blocked vehicle-lane-seconds to span the range of incident exposure, with three trials each) and 15 grid runs of the component variants, because it tests the predictive model under otherwise identical control rather than ranking controllers. Over the nine paired runs, SF-MPC has a 72.6~s (31\%) higher traveler ATT than GNN-MPC [66.2, 79.0] (run-level intervals; with three scenarios, clustered intervals are not informative), completes 157 fewer trips and leaves 139 more vehicles never inserted, close to the 168 more left by GNN-MPC (No Upstream) on the same runs (Supplementary Material, Sec.~S9-D). Upstream observation alone therefore does not reproduce the performance obtained with the learned dynamics: the upstream-halt term predicted by SF-MPC barely responds to the candidate phases, so that, as without the upstream representation in Section~\ref{subsec:observation}, its plans leave more of the demand at the network boundary. SF-MPC-Ext, which adds arrivals into the upstream zones, storage limits on the actual window lengths and spillback, has a 78.2~s (34\%) higher traveler ATT than GNN-MPC [68.9, 87.5], higher in all nine runs; relative to SF-MPC, it leaves fewer vehicles uninserted, with unresolved differences in traveler ATT and completed trips. Adding physical detail to the analytical predictor thus does not by itself improve control, although the contributions of the three extensions are not identified. On the grid, the substitution raises the traveler ATT consistently but by only 3.3~s [3.0, 3.7].

Offline accuracy is consistent with, but does not establish, this control value (Supplementary Material, Sec.~S9-E). The learned model predicts the lane queue more accurately than both analytical models on both networks and over the whole horizon: on Ingolstadt, its mean absolute queue error at the end of the prediction horizon is 0.170 vehicles per lane, against 0.206 and 0.202 for the analytical predictors and 0.199 for persistence, and on the grid, its one-step error is 0.027 against 0.059 for store-and-forward. The queue and upstream-halt terms account for about 97\% of the planning cost (median); the analytical predictors are, however, more accurate on the upstream halt, persistence remains competitive on the upstream halt and the waiting time, and errors grow on incident-affected lanes (Supplementary Material, Sec.~S5). Since offline errors are measured under the recorded phases, an action-ranking test asks instead whether each predictor identifies the best of the admissible phases of one intersection, simulated from the same recorded state (Supplementary Material, Sec.~S9-F). In the predefined class of intersections with mostly short lanes and upstream zones, the learned model identifies the best phase in 74\% of the 93 states, against 60\% for both analytical models (14 percentage points, [2, 26] under resampling of runs and [0, 28] under resampling of intersections), whereas the pooled difference over all 150 states is not resolved. Over all 150 states, the upstream-halt term spans on average about a third of the range of the predicted cost across the candidates for the learned model, against 9\% for SF-MPC, whose term is identical across candidates in more than half of the states; removing the term changes the phase selected by the learned model in 15\% of the states, against 3\% for both analytical predictors. The closed-loop disadvantage of SF-MPC is likewise concentrated at the short-lane entries with upstream zones (Supplementary Material, Sec.~S7-E). Within the three evaluated scenarios, replacing the learned dynamics with either analytical predictor thus substantially degrades closed-loop performance; the offline and ranking analyses are consistent with this result without fully explaining it.

\subsection{Component Ablations}
\label{subsec:ablation}

\begin{figure*}[t]
\centering
\includegraphics[width=\textwidth]{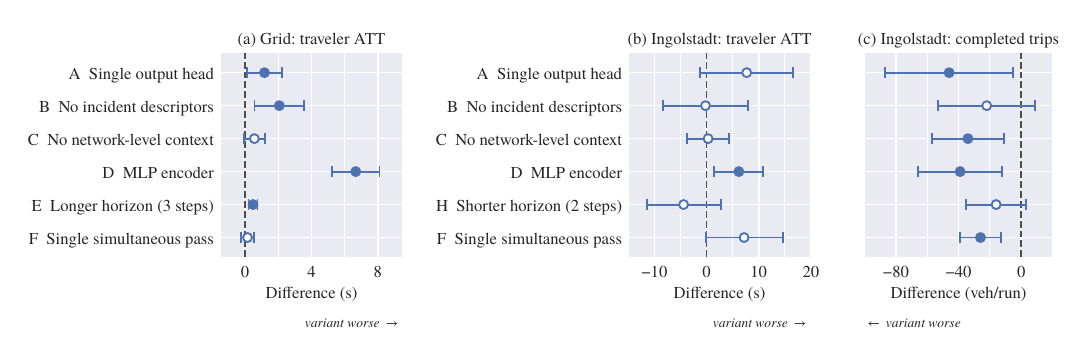}
\caption{Component ablations: paired difference variant minus full GNN-MPC, mean with 95\% Student-$t$ interval; filled markers: interval excludes zero (exploratory). (a)~Grid, traveler ATT, 15 pairs; (b)~Ingolstadt, traveler ATT, and (c)~completed trips, nine pairs. A:~single output head; B:~no incident descriptors; C:~no network-level context in the lane model; D:~MLP encoder; E/H:~longer/shorter horizon; F:~single simultaneous pass. Definitions in Supplementary Material, Sec.~S10.}
\label{fig:ablation}
\end{figure*}

Fig.~\ref{fig:ablation} shows how sensitive the closed-loop result is to the main design choices of the learned model and the planner. Each row is a variant of GNN-MPC with one change, and each marker is its mean paired difference from the full configuration with the 95\% interval: a marker to the right of zero in travel time, or to the left of zero in completed trips, means that the variant performs worse, and filled markers indicate intervals that exclude zero. The ablations characterize the sensitivity of the integrated configuration rather than the independent contribution of each component, since some modifications entail further changes; with 15 grid and nine Ingolstadt pairs and many comparisons, intervals that exclude zero are exploratory. Replacing the LSTM encoder (D) is the clearest delay sensitivity, raising the traveler ATT by 6.7~s [5.2, 8.1] on the grid and by 6.2~s [1.5, 10.8] on Ingolstadt, with 39 fewer completed trips. On Ingolstadt, removing the network-level predictions from the lane-model inputs (C) and using a single output head (A) reduce completed trips by 34 and 46 without resolved traveler-ATT differences; both entail retraining the lane model, and a configuration that additionally removes the network-level inflow forecast does not reduce completed trips further, so no monotonic benefit of every network-level prediction is established. Withholding the incident descriptors (B) raises the grid traveler ATT by 2.1~s [0.6, 3.5], mostly in E3, with no resolved effect on Ingolstadt. The tested horizon changes (E, H) do not improve on the main configuration, and a single simultaneous pass without acceptance test (F) costs 26 completed trips on Ingolstadt, a sensitivity to the update procedure as a whole. Several changes in completed trips are not resolved in traveler ATT, which supports reporting throughput alongside travel time; none of the ablations shows that a single component is necessary or sufficient for the advantage of Sections~\ref{subsec:results_overall}--\ref{subsec:sf}.

\subsection{Computation and Planner Behavior}
\label{subsec:computation}

Both networks were timed with the same protocol: per decision, on the same machine with a single consumer GPU, with the models loaded once and five warm-up states followed by 50 recorded states (Table~\ref{tab:computation}). Deterministic execution is the configuration of all closed-loop runs; non-deterministic execution is the faster default GPU mode that a deployment would use. On Ingolstadt, the main configuration meets the 10~s control interval under non-deterministic execution, also at its slowest decision (mean 6.5~s, maximum 8.9~s), and averages 11.1~s under deterministic execution; both modes perform identical search work. On the grid, the mean time per decision is about the length of the interval (9.9~s non-deterministic, 10.9~s deterministic), and its slowest decisions take up to 14.7~s; both modes yield identical plans in all 50 states. The shorter-horizon variant H is about four times faster on Ingolstadt (2.7~s deterministic, 1.6~s non-deterministic), without a measurable loss in control quality (Section~\ref{subsec:ablation}). The simulation assumes each plan to be available at the decision time, so computation delay is not modeled (Supplementary Material, Sec.~S11).

\begin{table}[t]
\centering
\caption{Planning time of GNN-MPC per decision (s) on 50 recorded states after five warm-up states, mean / p95 / observed maximum, on an NVIDIA RTX~5080. Deterministic: execution mode of all closed-loop runs; non-deterministic: default GPU execution. H: shorter horizons ($n_u = n_y = 2$).}
\label{tab:computation}
\scriptsize
\setlength{\tabcolsep}{2pt}
\begin{tabular}{lrr}
\toprule
 & \textbf{Grid} & \textbf{Ingolstadt} \\
\midrule
Main configuration, deterministic & 10.9 / 15.8 / 15.9 & 11.1 / 15.0 / 15.1 \\
Main configuration, non-deterministic & 9.9 / 14.2 / 14.7 & 6.5 / 8.8 / 8.9 \\
H (2 / 2), deterministic & -- & 2.7 / 3.6 / 3.7 \\
H (2 / 2), non-deterministic & -- & 1.6 / 2.1 / 2.1 \\
\bottomrule
\end{tabular}
\end{table}

Over all closed-loop decisions, the planner uses 2.1 (grid) and 2.3 (Ingolstadt) passes on average and at most five, and the accepted predicted cost was non-increasing in every decision, consistent with Proposition~\ref{prop:monotone_descent}(i). Most decisions ended at the relative-improvement tolerance, so the candidate-set equilibrium of part~(iii) was reached in only 0.4--3.4\% of them. These measurements show neither real-time deployability in closed loop nor closed-loop stability.

\section{Discussion and Conclusion}
\label{sec:conclusion}

Under the evaluated incident conditions, GNN-MPC achieved the lowest traveler ATT and the most completed trips on both networks. The controlled comparisons on Ingolstadt indicate that representing upstream demand mainly changes how much of the demand is admitted, whereas the predictive model also determines how much of it is served: with comparable observation coverage and nearly identical admission, GNN-MPC had a 13\% lower traveler ATT than MaxPressure-U, and replacing its learned dynamics with an analytical predictor raised the traveler ATT by 31--34\% on Ingolstadt and by 3~s on the grid. These comparisons support complementary roles for observation coverage and predictive-model choice, without isolating their interaction or establishing that the benefits arise specifically from incidents.

For predictive control of networked systems with partial measurements, the results point to three interdependent design considerations: which states the controller observes, how its model predicts their response to alternative actions, and how the predictions coordinate interacting decisions. Under the same traffic measurements and planning, the choice of predictive model materially changed admission and completed service, an effect that offline prediction errors alone do not establish. Since controllers can also differ in how much of the demand they admit, evaluating them on the whole scheduled demand, together with admission and throughput, gives a more complete account of their service.

Several limitations qualify these conclusions. The evidence covers two networks and a single demand profile on Ingolstadt, and the predictor substitutions and ablations nine and 15 paired runs; each learning-based method is represented by one trained model, and the training condition of each RL baseline was selected on the evaluation runs, which favors the baselines. Incident information is assumed to be available immediately (Assumption~\ref{ass:incident}) and is given to GNN-MPC but not to the RL and pressure-based baselines, and without incident-free runs on Ingolstadt, incident effects cannot be separated from network load. The learned dynamics are trained per network, mainly on data collected under MaxPressure, and the analytical comparison covers two store-and-forward formulations only. The main planning configuration meets the 10~s interval on Ingolstadt under non-deterministic execution and takes about the length of the interval on the grid, whereas the closed-loop runs use deterministic execution, and computation delay is not modeled; the guarantees concern the predicted cost at one decision step and do not establish closed-loop stability.

Future work will reduce the planning cost further through neighborhood-restricted rollouts, model computation and detection delays in closed loop, train the dynamics on data from multiple controllers, and test transfer to unseen networks and demand conditions. Under the evaluated incident conditions, combining upstream-demand representation, learned action-conditioned dynamics and coordinated receding-horizon planning improved the service of the scheduled demand, illustrating that observing hidden demand and predicting its response to control are complementary requirements of network-level predictive control.

\appendices
\section{Proof of Proposition~\ref{prop:monotone_descent}}
\label{app:proof}
Throughout, $J_t$ is the fixed deterministic function of Proposition~\ref{prop:monotone_descent}.
(i) If the update at step $k$ is accepted, $J_t(U^{(k+1)})=J_t(\tilde{U}^{(k)})<J_t(U^{(k)})-\varepsilon$ by~\eqref{eq:acceptance}; otherwise, $U^{(k+1)}=U^{(k)}$.
(ii) The set $\mathcal{U}$ is finite, so $J_{\min}=\min_{\mathcal{U}}J_t$ exists and $J_t(U^{(k)})\geq J_{\min}$ for all $k$. If $\varepsilon>0$, each accepted update lowers $J_t$ by more than $\varepsilon$ by~(i), which can occur at most $(J_t(U^{(0)})-J_{\min})/\varepsilon$ times. If $\varepsilon=0$, $J_t$ strictly decreases at each accepted update, so $U^{(0)}$ and the plans reached by accepted updates are pairwise distinct elements of $\mathcal{U}$; hence, there are at most $|\mathcal{U}|-1$ accepted updates. Every pass either contains an accepted update or does not, and accepted updates are finite in number, so some pass contains none.
(iii) During a pass without an accepted update, the plan remains $U^\star$. For each intersection $v$, rejection in~\eqref{eq:acceptance} gives $J_t(\tilde{U}_v,U^\star_{-v})\geq J_t(U^\star)-\varepsilon$, and~\eqref{eq:approx_br} gives $J_t(U_v,U^\star_{-v})\geq J_t(\tilde{U}_v,U^\star_{-v})$ for all $U_v\in\mathcal{S}_v$; together, they yield~\eqref{eq:eps_equilibrium}. With $\varepsilon=0$ and $\mathcal{S}_v=\mathcal{U}_v^{n_u}$, this is the definition of a Nash equilibrium. \hfill$\blacksquare$
\bibliographystyle{IEEEtran}
\bibliography{cas-refs}
\clearpage
\includepdf[pages=-]{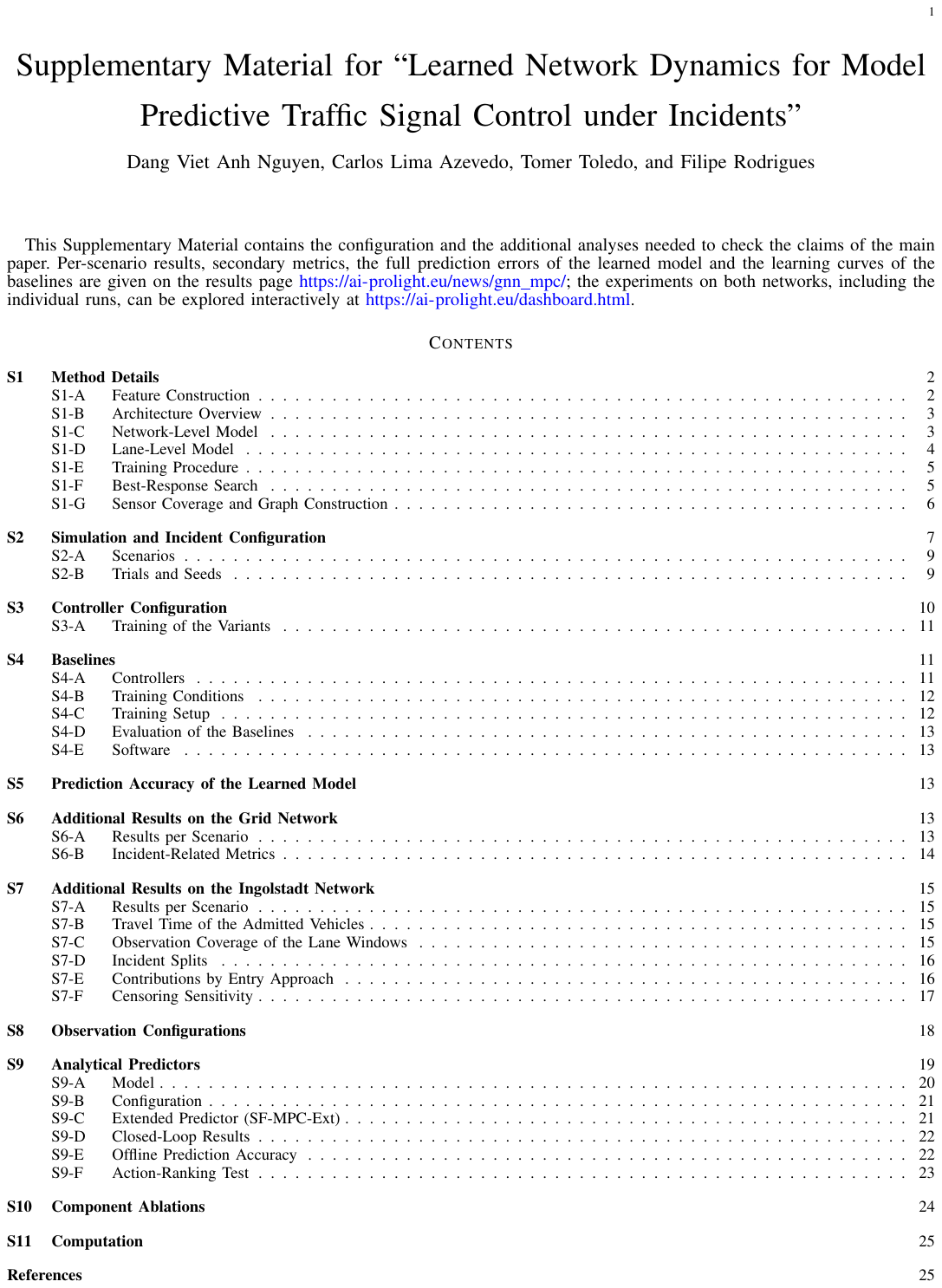}
\end{document}